%% file: paper.tex
\documentclass{article}

\usepackage{balance}       
\usepackage{graphics}      
\usepackage[utf8]{inputenc}
\usepackage[T1]{fontenc}   
\usepackage[main=american]{babel}
\usepackage[babel=true]{csquotes}
\usepackage[round]{natbib}
\usepackage{enumitem}
\usepackage{txfonts}
\usepackage{xspace}
\usepackage{mathptmx}
\usepackage{color}
\usepackage{booktabs}
\usepackage{textcomp}
\usepackage{suffix}
\usepackage{xcolor}
\usepackage{subcaption}
\usepackage{tabularx}
\usepackage[multiple]{footmisc}
\usepackage[pdflang={en-GB},pdftex]{hyperref}

\usepackage{microtype}        
\usepackage{ccicons}          
\usepackage{todonotes}

\def \plaintitle {The Impact of Text Presentation\\on Translator Performance}

\definecolor{linkColor}{RGB}{6,125,233}
\hypersetup{%
    pdftitle={\plaintitle},
    pdfauthor={},
    pdfkeywords={},
    pdfdisplaydoctitle=true, 
    bookmarksnumbered,
    pdfstartview={FitH},
    colorlinks,
    citecolor=black,
    filecolor=black,
    linkcolor=black,
    urlcolor=linkColor,
    breaklinks=true,
    hypertexnames=false
}

\input{commands}

\newcommand\Affil[1]{\textsuperscript{#1}}

\begin{document}

\title{\plaintitle}
\author{
    Samuel Läubli\,\Affil{a}\footnote{Work carried out at Lilt, Inc.} \quad 
    Patrick Simianer\,\Affil{b} \quad 
    Joern Wuebker\,\Affil{b}\\
    Geza Kovacs\,\Affil{b} \quad 
    Rico Sennrich\,\Affil{a,c} \quad 
    Spence Green\,\Affil{b}\\[2ex]
    \Affil{a}\,University of Zurich \quad 
    \Affil{b}\,Lilt, Inc. \quad 
    \Affil{c}\,University of Edinburgh\\
}
\date{}

\maketitle

\begin{abstract}
Widely used computer-aided translation (CAT) tools divide documents into segments such as sentences and arrange them in a side-by-side, spreadsheet-like view. We present the first controlled evaluation of these design choices on translator performance, measuring speed and accuracy in three experimental text processing tasks. We find significant evidence that sentence-by-sentence presentation enables faster text reproduction and within-sentence error identification compared to unsegmented text, and that a top-and-bottom arrangement of source and target sentences enables faster text reproduction compared to a side-by-side arrangement. For revision, on the other hand, our results suggest that presenting unsegmented text results in the highest accuracy and time efficiency. Our findings have direct implications for best practices in designing CAT tools.
\end{abstract}

\section{Introduction}

Research into CAT tool adoption among professional translators shows that poor usability is a major reason for resistance \citep{LeBlanc2013,OBrien2017}. The sentence-by-sentence presentation of texts, for example, was criticised by translators for creating an \enquote{obstructed view of the text, which in turn disrupts the workflow} \citep{OBrien2017}. However, the impact of poor usability on translator performance has rarely been tested empirically, and since the motivation for using CAT tools is primarily economic -- saving time by leveraging translation suggestions rather than translating from scratch -- the design of these tools is unlikely to change until measurements show that alternative designs speed translators up or cause them to make fewer mistakes.

In this article, we test the impact of text presentation on translator performance in three text processing tasks. Our motivation is two-fold: controlled experiments show that text presentation affects reading performance \citep{Hornbaek2001,YuMiller2010}, and that access to linguistic context affects judgement of translation quality \citep{Laeubli2018}; qualitative research finds that text presentation in CAT tools is irritating \citep{LeBlanc2013,OBrien2017}, and some translators interviewed in this work think that working with continuous rather than segmented text would help solve some \enquote{really hard trouble} in their daily work (\Section{Exp2Survey}).
We hypothesise that the empirical findings from experiments on reading and quality evaluation, two inherent activities when working with CAT tools, will carry over to computer-aided translation and substantiate concerns expressed by professional translators.

    
    
    
    

\begin{figure}
    \makebox[\linewidth][c]{
        \begin{subfigure}[t]{0.68\textwidth}
            \centering
            \includegraphics[width=\textwidth,trim={4cm 4cm 4cm 4cm}, clip]{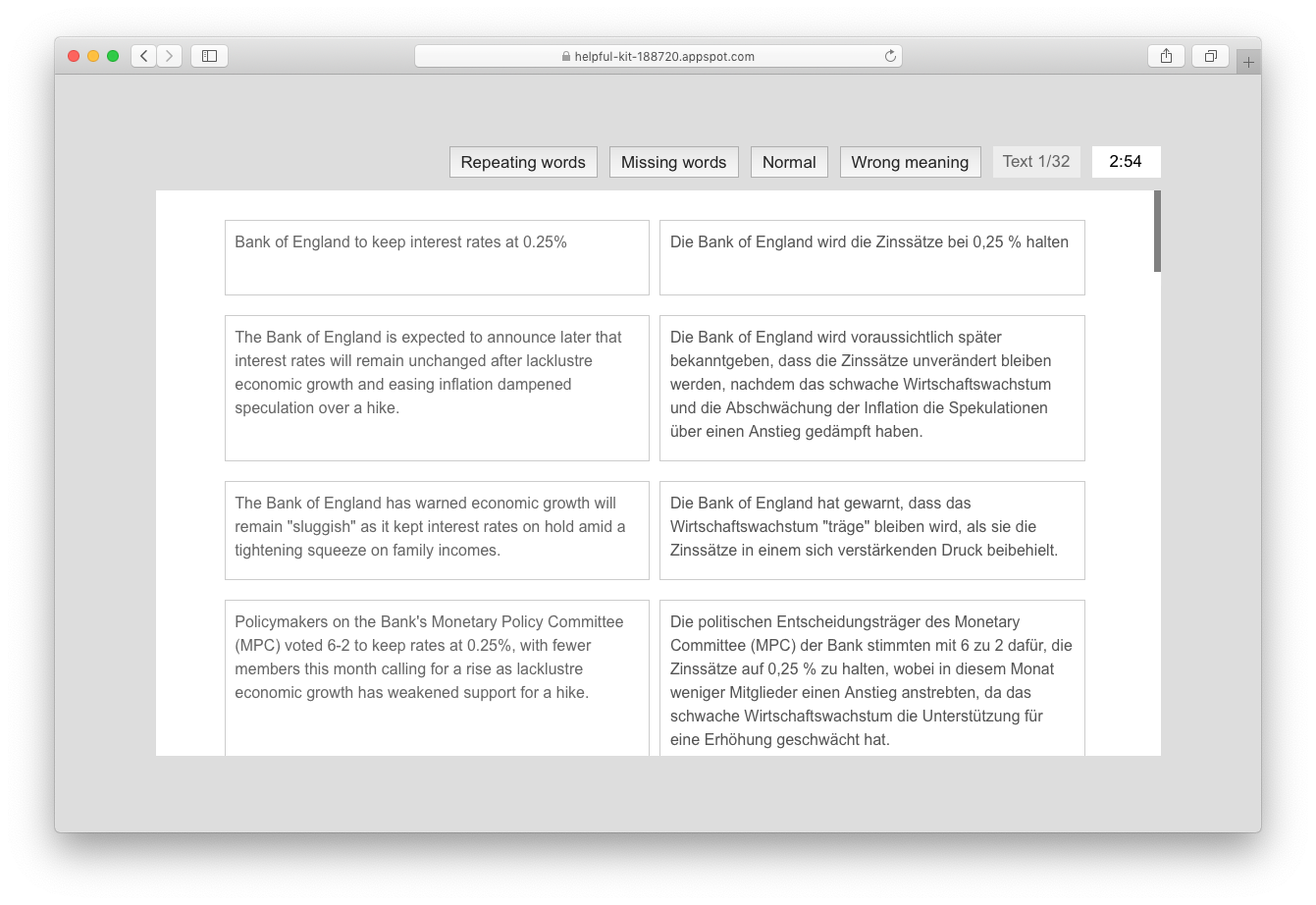}
            \caption{Sentence, left--right (SL)}
            \label{fig:Exp2ScreenSL}
            \bigskip
        \end{subfigure}
        \quad~~
        \begin{subfigure}[t]{0.68\textwidth}
            \centering
            \includegraphics[width=\textwidth,trim={4cm 4cm 4cm 4cm}, clip]{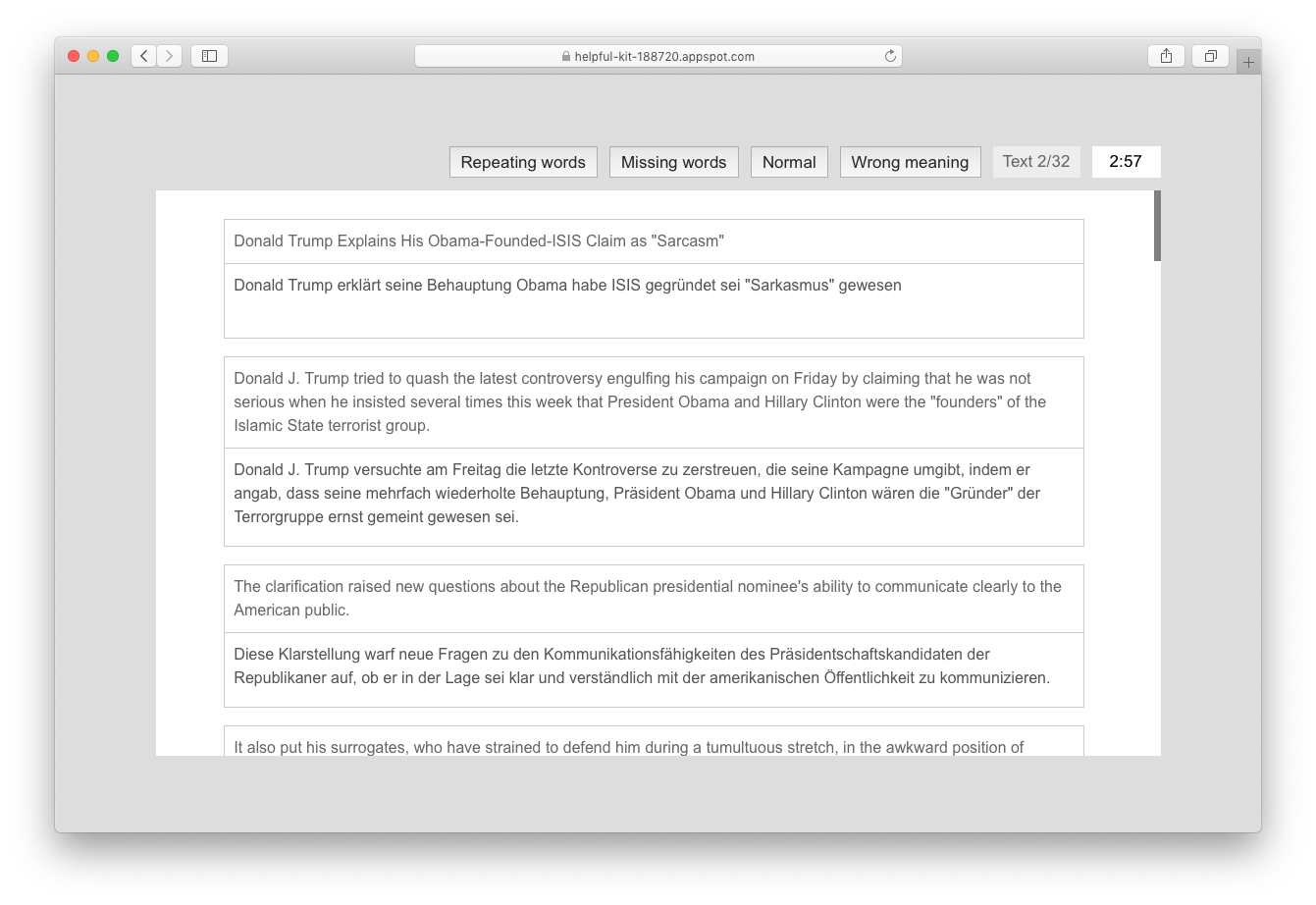}
            \caption{Sentence, top--bottom (ST)}
            \label{fig:Exp2ScreenST}
            \bigskip
        \end{subfigure}
    }
    \makebox[\linewidth][c]{
    \begin{subfigure}[t]{0.68\textwidth}
        \centering
        \includegraphics[width=\textwidth,trim={4cm 4cm 4cm 4cm}, clip]{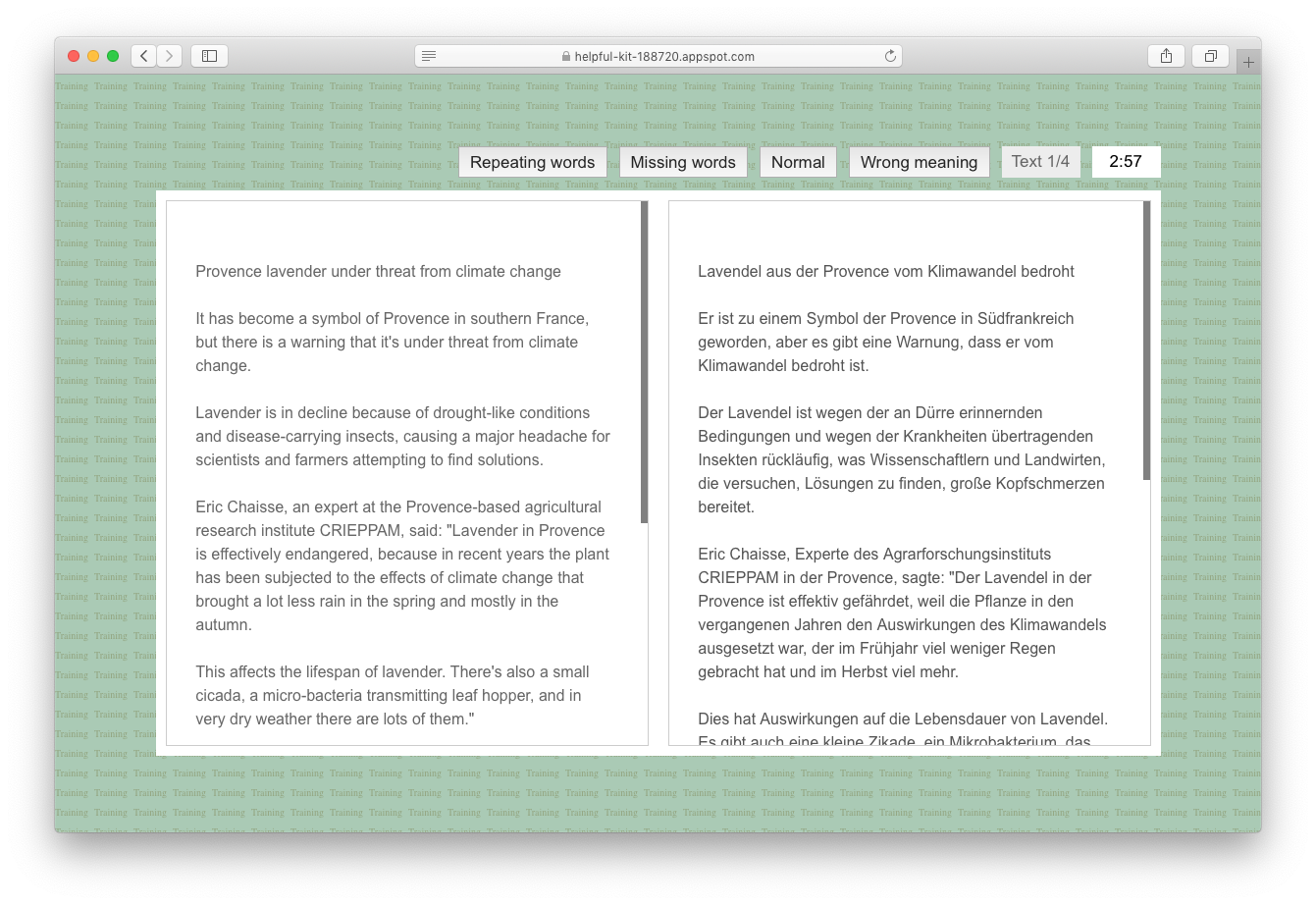}
        \caption{Document, left--right (DL)}
        \label{fig:Exp2ScreenDL}
    \end{subfigure}
    \quad~~
    \begin{subfigure}[t]{0.68\textwidth}
        \centering
        \includegraphics[width=\textwidth,trim={4cm 4cm 4cm 4cm}, clip]{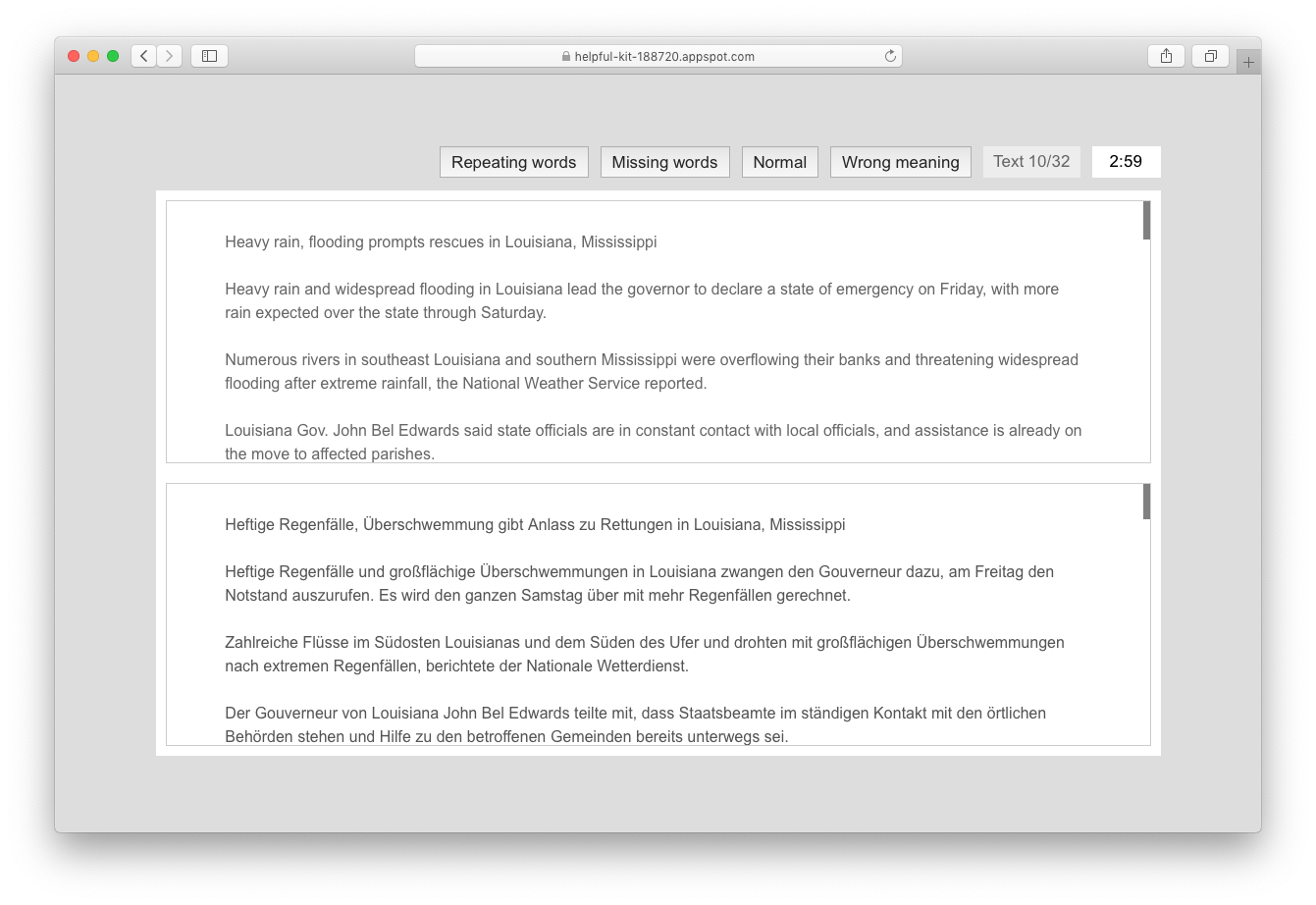}
        \caption{Document, top--bottom (DT)}
        \label{fig:Exp2ScreenDT}
    \end{subfigure}
    }
    \captionsetup{width=1.44\textwidth}
    \caption[Experimental UIs.]{UI configurations evaluated in this study. We test the efficacy of sentence segmentation (S) \vs full document presentation (D) and left--right (L) \vs top--bottom (T) orientation.}
    \label{fig:Exp2Screen}
\end{figure}

Our investigation is focused on two aspects of text presentation: segmentation and orientation. Widely used CAT tools\footnote{Examples include Across, MemoQ, and Trados Studio \citep{Schneider2018}.} segment texts into sentences and present them in a side-by-side, spreadsheet like view (\Figure{Exp2ScreenSL}). Sentence segmentation is a natural choice from a technical perspective because backend technologies that provide translation suggestions, such as translation memory (TM) and machine translation (MT) systems, operate at the level of sentences,\footnote{TMs can be configured to operate at the level of paragraphs, but since retrieval rates are lower, sentence-level segmentation is more common. However, context matches consider some super-sentential information, such as the previous segment. Document-level MT \citep[\eg,][]{JunczysDowmunt2019} is not available commercially at the time of writing.}
but translators consider document-level discourse. When sentences are placed in separate boxes, inter-sentential references, such as a pronoun and its antecedent, are placed further apart, so a user interface (UI) that presents continuous text may be more suitable for spotting errors related to textual cohesion (compare Figures~\ref{fig:Exp2ScreenSL} and~\ref{fig:Exp2ScreenDL}). Similarly, the distance between a word in the source and its suggested translation in the target text is larger when sentences are shown side-by-side (\Figure{Exp2ScreenSL}) compared to a top-and-bottom configuration (\Figure{Exp2ScreenST}). \citet{Green2014PTM} conjecture that the latter would reduce gaze shift, the time it takes translators to realign their line of sight to the relevant segment, and we assume that a UI that eases visual orientation will lead to faster and more accurate translation.

Measuring translator speed and accuracy in controlled translation experiments is challenging: a subject cannot be exposed to the same translation in different conditions due to repetition priming, and translation quality is difficult to define and measure \citep{House2013,Green2013}. To control for confounding variables, we focus on specific activities that can be relevant when working with CAT tools -- text reproduction, within-sentence error identification, and document-level revision -- and design our experimental tasks such that accuracy\footnote{Throughout this article, we use the term accuracy rather than quality to emphasise that we focus on specific linguistic phenomena that are categorisable as correct or incorrect with no or minimal ambiguity.} can be measured with minimal ambiguity. In the revision task, for example, we insert errors into human translations that are unambiguously wrong, and measure whether and how quickly subjects correct these errors within the different UIs.

We review related work, and previous studies that use similar means of experimental control, in \Section{Exp2Background}. In \Section{Exp2Survey}, we report on semi-structured interviews with professional translators to assess the practical viability of design changes related to text presentation in CAT tools. Our experimental design and results are presented in Sections~\ref{sec:Exp2Methods} and~\ref{sec:Exp2Results}, respectively. We discuss implications for best practices in designing CAT tools, alongside limitations of our experimental design and results, in \Section{Exp2Discussion}, and draw conclusions in \Section{Exp2Conclusions}.

\section{Background}
\label{sec:Exp2Background}

Our interaction with computers, machines that carry out mathematical operations, is mediated by UIs. When we work with graphical UIs, we tend to forget that everything we see is the result of a design process: the position, colour and size of any button and text box are not determined by chance, but by design decisions actively made by people in charge \citep{Norman1988}. As it is difficult to test every option with the intended audience, designers base some or all of these decisions on conventions and assumptions. A convention in the UI of a program running on the Windows operating system, for example, is to place a small red button with an \enquote{x} symbol, whose on-click behaviour is to terminate the program, in the upper right corner. The assumption is that users will be familiar with this convention and thus know how to terminate the program, but if there are no conventions or if conventions are considered suboptimal, metaphors are a powerful tool for designers. For instance, the adoption of personal computers soared after the metaphor \textsc{Computer is a Desktop} replaced the conception that a computer is a programming environment \citep{Saffer2005}. 


\subsection{Text and Document Visualisation}
\label{sec:Exp2BackgroundVisualisation}


Text editing has been a fundamental task supported by modern computers since their inception \citep{EngelbartEnglish1968}. Early text editors were referred to as line editors because, mostly due to hardware constraints, users were required to select, manipulate, and then display individual lines of a document in separate steps; manipulation and document display did not occur simultaneously. \citet{Shneiderman1983} promoted display editors: whereas \enquote{the one-line-at-a-time view offered by line editors is like seeing the world through a narrow cardboard tube}, a display editor always shows a document in its final form and \enquote{enables viewing each sentence in context and simplifies reading and scanning}. The visualisation of full documents has a direct impact on productivity: display editors were shown to double text editing speed compared to line editors \citep{Roberts1980,RobertsMoran1982}.

While the \enquote{what you see is what you get} (WYSIWYG) principle seen in display editors has long become the standard in word processing software we use in our everyday life, adjustments in text presentation have further improved UIs for text editing. \citet{Hornbaek2001} investigate if two alternatives to a regular (referred to as linear) UI improve reading speed and comprehension of electronic documents: a fisheye UI that shrinks certain parts of the document below readable size, which can be made readable by clicking on them; and an overview+detail UI that displays a miniaturised version of the document in a sidebar (the overview pane) that can be clicked to quickly move the main pane (referred to as the detail pane) to a desired section. A controlled experiment with 20 subjects finds that while the fisheye UI improves reading speed, the overview+detail UI improves reading comprehension and achieves the highest satisfaction among subjects, ten of which \enquote{mention the overview of the documents structure and titles as an important reason} (\ibid). While we do not use an overview pane in our experimental interfaces, we observe that much of a document's structure is lost in the sentence-level UIs of widely used CAT tools (\Figure{Exp2ScreenSL}), while a UI that presents unsegmented text retains structural cues such as titles and paragraphs (\eg, \Figure{Exp2ScreenDL}) or lists of various types. \citeg{YuMiller2010} Jenga format is a compromise between the two: it separates paragraphs into sentences, but, in contrast to CAT tools, only adds vertical space while the horizontal position of each sentence remains unchanged. A user study with 30 subjects finds that presenting texts in this format significantly enhances web page readability (\ibid).

\subsection{Text and Document Visualisation in CAT Tools}
\label{sec:Exp2BackgroundTranslationTechnology}

Context is also vital for translators to produce high-quality translations, yet this context is often narrow in CAT tools. One reason is that, in contrast to regular word processors, the UI of a CAT tool needs to accommodate two documents -- the (generally uneditable) source document and its translation, the target document -- and additional panes to display translation suggestions. All of these elements compete for space on the translator's screen, and while the size of these elements is typically configurable, showing more translation suggestions at once, for example, will necessarily decrease the number of source and target sentences that can be shown without scrolling. Another reason is that the source and target documents are rendered as a table where each sentence is placed in a separate cell. If a sentence does not use the full width of a cell, or if either the source or the target sentence uses more lines than its counterpart, some of the space remains blank, which further limits the number of sentences that can be viewed without scrolling. A UI that shows continuous text can accommodate more text -- and thus more context around the sentence being translated (compare Figures~\ref{fig:Exp2ScreenSL} and~\ref{fig:Exp2ScreenDL}).

The fact that widely-used CAT tools visualise documents as tables rather than continuous text implies a common motivation among manufacturers, and the question that arises is whether this motivation is rooted in ergonomic considerations. From a user's perspective, the \textsc{Document is a Table} metaphor seems less intuitive than the \textsc{Document is a Series of Pages} metaphor used in applications like Microsoft Word, which, despite not offering translation functionality, is used for MT post-editing by 38\percent of professional translators \citep{MoorkensOBrien2017}. Translation process research finds that the sentence-by-sentence presentation in CAT tools \enquote{creates an unnaturally strong focus on the sentence} that reduces the number of changes made to sentence structure in translations \citep{Dragsted2006}, and ethnographic studies as well as surveys with professional translators conclude that the segmented view of documents is problematic \citep[\eg,][]{LeBlanc2013,OBrien2017}.


While we have been unable to find published information that motivates the use of sentence segmentation by commercial CAT tool providers, a review of academic research suggests that design choices on document visualisation are not based on empirical investigation. \citet{Kay1980} suggests incorporating simple translation functionality into word processors. He theorises that this editor would be \enquote{divided into two windows. The text to be translated appears in the upper window and the translation will be composed in the bottom one}. This suggestion -- a document-level UI with top--bottom orientation -- was later implemented in TransType: \citet{Langlais2001} \enquote{tried to display the text and its translation side by side but \textit{it seems} that a synchronized display of the original text and its translation one over the other is better} (emphasis added). Translog-II \citep{Carl2012}, a tool widely used in translation process research, also arranges the unsegmented source and target document in a top--bottom configuration. \citet{Green2014PTM} present a UI that uses a top-and-bottom arrangement of source and target sentences instead of unsegmented source and target documents. The authors ground this design choice in the observation that translators spend up to 20\percent of their time reading when translating a document \citep{Carl2010}, and argue that their

\begin{quote}
    UI is based on a single-column layout so that the text appears as it would in a document. Sentences are offset from one another primarily because current MT systems process input at the sentence-level. We interleave target-text typing boxes with the source input to minimize gaze shift between source and target. Contrast this with a two-column layout in which the source and target focus positions are nearly always separated by the width of a column.
\end{quote}

\citeg{Green2014PTM} investigation is focused on interaction features and does not assess the impact of top--bottom orientation. The CAT tool prototype evaluated by \citet{Coppers2018} also uses the design proposed by \citet{Green2014PTM}, but the authors do not evaluate it against a UI that uses left--right orientation, a gap we fill with the experiment presented in this article.

\subsection{Understanding Translator Performance}
\label{sec:Exp2BackgroundPerformance}

Our aim is to assess the impact of text presentation on translator performance, and a fundamental question in translation experiments is how translator performance should be defined and measured. Some experimental designs maximise external validity: they measure temporal effort and/or the quality of products under realistic working conditions, the goal being that results will reflect the \enquote{truth in real life} \citep[\eg,][]{Federico2012}. Apart from resource-related challenges such as high cost (\eg, because subjects should be professional translators rather than students), such experimental designs limit control of extraneous variables (\eg, because the user-defined settings in a CAT tool cannot be standardised when subjects use their own workstation) and insights into why a particular result was obtained (\eg, whether slower subjects spend more time on reading or writing). Moreover, realistic working conditions may not be achievable in the context of fundamental research not only because subjects will necessarily be unfamiliar with the research prototypes to be tested, but also because prototypes will typically not implement all of the functionality available in commercial products.

For some or all of these reasons, other experimental designs in translation research maximise internal validity. 
In ensuring that results will reflect \enquote{the truth in the study}, such designs may involve resources and procedures that deviate from realistic working conditions for better control (\eg, control for screen size by having all subjects work on a standardised workstation in a lab) or finer-grained measurements (\eg, how much time subjects spend reading and writing). The investigation of \citet{Krings1994,Krings2001},\footnote{We reference page numbers in the English translation of \citeg{Krings1994} habilitation thesis \citep{Krings2001} due to better availability and accessibility.} for example, aims at gaining an understanding of how translation processes change as translators post-edit MT rather than translate from scratch.\footnote{Even if his study is best known for the finding that the temporal effort for translation from scratch and MT post-editing (in 1994) is roughly the same \citep[p.~552]{Krings2001}, which, as such, can also be tested with an extrinsic design \citep{Federico2012}.}
\citeauthor{Krings2001} asks subjects to Think Aloud \citep{EricssonSimon1984} as, using pen and paper, they translate or post-edit, the latter without access to the source text in one task of the experiment. Although very different from a translator's regular working conditions, this setup allows the author to elaborate and quantify the relative distribution of sub-processes, such as target text monitoring or writing, in translation from scratch and post-editing. The use of Think Aloud protocols is known to impact translation speed \citep{Jakobsen2003}, and other data collection methods such as key-logging and eye-tracking likewise pose challenges to external validity \citep[\eg,][]{OBrien2009}; but while results like time measurements from such experiments may not be directly transferable to real-life situations, conclusions drawn from comparing measurements between experimental conditions may well be. With respect to \citet{Krings2001}: while it may not hold that 42.5\percent and 43.5\percent of the processes in translation from scratch and post-editing, respectively, relate to target text production under normal working conditions (\ibid, p.~314), it is plausible that the difference will also be small under normal working conditions since the aggravating circumstances were the same in both tasks of the experiment.

Since no commercial CAT tool implements all of the UIs we test in our experiment,\footnote{On the contrary, we are not aware of any CAT tool in wide use that implements a document-level UI.} the use of prototypes is inevitable, and our goal cannot be to predict how text presentation will affect translation under real-life working conditions with commercial CAT tools that provide many more functions than these prototypes (\Section{Exp2MaterialsUIs}). Instead, we are interested if, and to what extent, the different UIs impact the speed and accuracy of professional translators when all but segmentation and orientation -- such as font size, spacing, etc. -- stays exactly the same. Our experimental design choices are guided by two principles aimed at maximising internal validity. First, we do not categorise translation processes, but define specific tasks for particular processes (\Section{Exp2Tasks}). To assess how the UIs affect reading, for instance, we do not ask subjects to translate a text and then try to identify in which parts of the translation sessions subjects were reading; we define a specific reading task (\Scan). Second, we define response variables that are measurable with no or minimal ambiguity. To assess if the UIs impact the number of typing errors, for example, we do not look for typing errors in freely written translations; we ask translators to reproduce a given text (\Copy) so we can calculate the number of typing errors exactly. The specifics of our experimental design are detailed in \Section{Exp2Methods}.

\section{Preliminary Feedback from Potential Users}
\label{sec:Exp2Survey}

To assess if design changes in text presentation are considered viable for practical use in CAT tools, we conduct semi-structured interviews with professional translators.

\subsection{Method}

We recruit eight professional translators (I1--8) from a multinational language services provider. They are recommended by a project manager (convenience sampling) and participate voluntarily; we do not offer compensation. The interviewees have been working as full-time translators between 0.5 and 13 years (mean=4.4) have between 0.5 and 9 years of experience using CAT tools (mean=3.3). We walk them through a series of closed and open-ended questions.

\subsection{Ideation}

After a short introduction, we ask the interviewees what the CAT tool of their dreams would look like.
We encourage interviewees to think bold and leave budgetary or technical limitations aside. They are given access to an online whiteboard for collaborative sketching with the interviewer, and are told that they are free to use it or not.

Although our primary goal is to elicit feedback on text presentation, we do not share any ideas or explicitly mention the topic at this point. We want to see if the interviewees initiate the topic themselves, and some indeed wish for functionality related to visual context. I8 states that an ideal editor \enquote{should look like a Word document but with some addiction \sic on a translation workbench.} A number of interviewees tell us that it is hard for them to translate without knowing what the source and target texts look like. I3 says that \enquote{we spend a lot of time thinking like \enquote{What, where does this come from?}}, and that \enquote{the ideal thing would be to have a visual context for the text, like you know where that segment is in the final page.} 
Similarly, I2 would include some sort of preview in the CAT tool of their dreams, and I8 would like to \enquote{see the whole text} during translation: \enquote{I know it sounds small, but it would be really useful.}

\subsection{Concept Testing}

\begin{figure}
    \centering
    \includegraphics[width=0.8\textwidth,trim=5cm 5cm 8cm 5cm,clip]{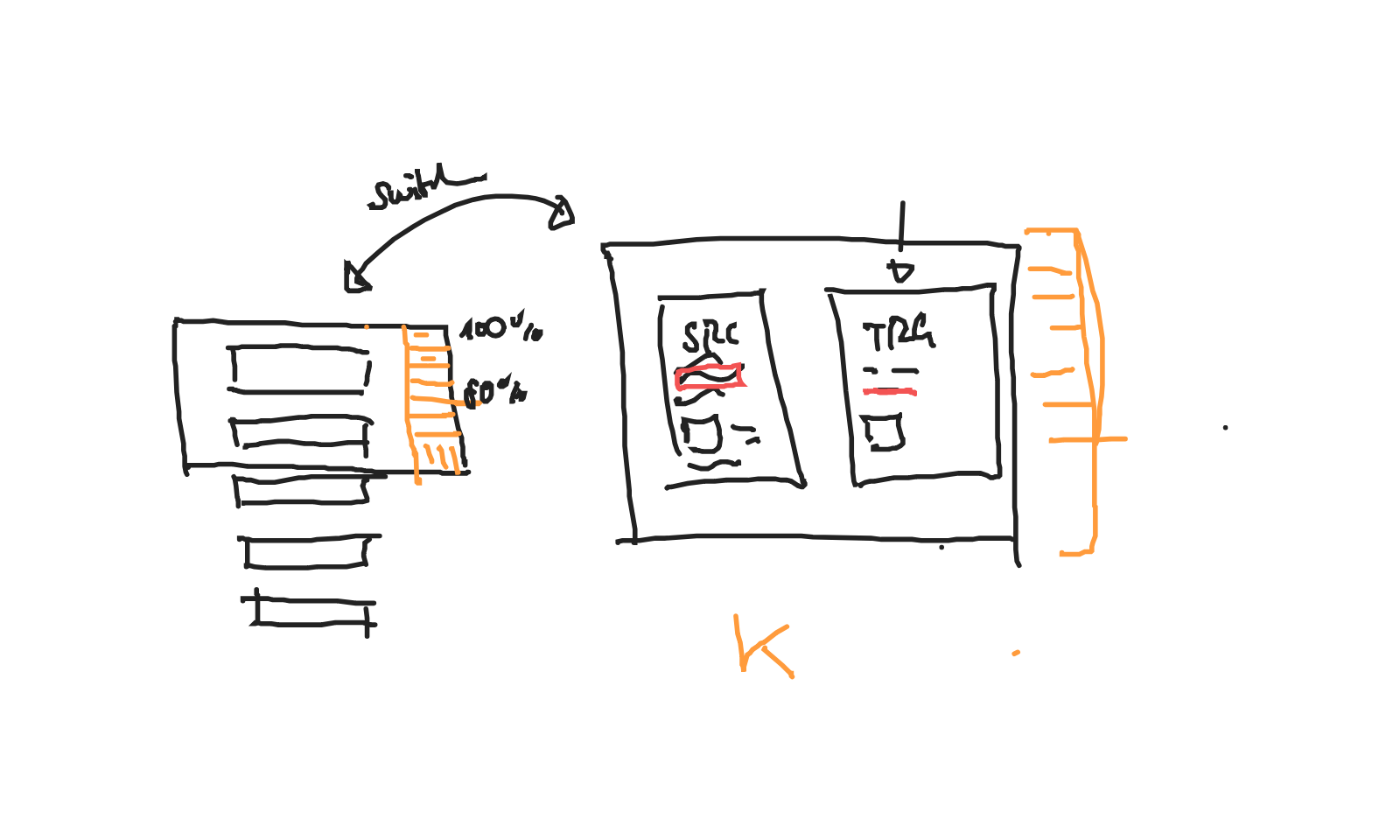}
    \caption{Collaborative sketching over the course of a semi-structured interview. The interviewee (I5) suggests to let users switch between sentence and document-level views.}
    \label{fig:Whiteboard}
\end{figure}

In the second part of the interview, we specifically focus on document-level editors. If interviewees do not bring up the topic themselves, we steer the conversation towards it by saying that we are \enquote{thinking about translation software that lets you focus more on documents as a whole rather than individual segments.}

We then illustrate the concept using the online whiteboard, as shown in \Figure{Whiteboard}. We start by sketching out how documents are split into sentences with current CAT tools, how each sentence is put into a box, and how only few boxes fit into the user’s screen and are thus visible at the same time. We then contrast this with a screen showing two entire pages: the left one containing the source text, the right one being empty. In that sense, our drawing resembles the \enquote{print layout} available in Microsoft Word or Google Docs, but with two parallel documents.

\subsubsection{Praise and Opportunities}

First impressions from seven interviewees are positive. When asked what they like about the concept, interviewees highlight the potential for better translation quality and ease of use. Some think that translations produced with such an editor \enquote{would be more true to the source} (I4) or \enquote{seem less as a translation than when you do the segment for segment thing instead} (I6). I8 says it would help solve some \enquote{really hard trouble} with the CAT tools they currently use: \enquote{Sometimes \etc when I’m going to send my files back to my clients they struggle to [put the translations into] the same layout.} Along the same lines, I7 likes that \enquote{you could see whether you’re translating a title, a subtitle \ldots}. 
I6 calls the concept \enquote{quite a big leap} and points out that it might ease the merging and splitting of segments:

\begin{quote}
As a translator in [a commercial CAT tool] I’m just looking at that one segment. Of course I know the segments that are around it, but I think, uhm, I would imagine that if you have the entire document as a thing that you’re translating you could maybe move some things from one sentence to another whereas in [the commercial CAT tool] you could never do that. Or you can, but it’s a lot of annoying stuff to merge segments together or split them. You’re basically tied to whatever the system has thought is a good way to segmentise the text. And that’s not how a person will read that text eventually, it’s sort of a mismatch between how the text will be used and how the translation is done.
\end{quote}

Talking about opportunities, I5 thinks that the concept is \enquote{like you’re using [a commercial CAT tool], but at the same time you get to see the final product which, you know, when you use [that CAT tool] you’re going blind until you’re done and you export it and you’re praying that it’s gonna be okay.} I3 adds that document-level editing could be helpful for reviewing in particular.

\subsubsection{Criticism and Limitations}

One interviewees's first reaction is negative, saying that \enquote{instinctively it feels like a step back for me} (I1). The concept reminds them of what translation was like before using CAT tools, when jumping back and forth with their eyes between two documents felt tiring; they find it helpful that CAT tools break texts down into \enquote{this nice long list}. I3 finds sentence-by-sentence presentation useful to \enquote{order yourself} and work bit by bit. In the same vein, I5 says that \enquote{you need segmentation sometimes to just better focus}; in line with I2 and I7, they figure that enabling translators to switch between segment and document-level views would probably be most effective (\Figure{Whiteboard}). Lack of orientation is thus the main reservation among interviewees. 
I6 mentions that even working with long sentences can be hard:

\begin{quote}
The one thing that I can compare this [document-level UI] with is whenever I have a large segment in [a commercial CAT tool], and when that happens I’m never happy because it’s annoying, you tend to use lose track of where you are because at some point there’s gonna be a mismatch between, uhm, oh in the English segment I’m like in the third row whereas in [other language] it’s a longer language so I’m probably gonna be on the fifth row already. It’s gonna be like not matching up correctly so you sort of check back all the time like \enquote{where was I?}.
\end{quote}

Some interviewees also note that document-level UIs would be unsuitable for certain text types, such as a list of keywords. I4 says that a document-level UI would only be helpful \enquote{if you’re dealing with a whole document like an article or a medical record or something as a CV}.

\section{Experimental Methods}
\label{sec:Exp2Methods}

We conduct a controlled experiment to empirically test the impact of text presentation on translator performance. We use a mixed factorial design and measure time and accuracy in three experimental tasks. The independent variables (factors) are UI segmentation (S: Sentence, D: Document), UI orientation (L: Left--Right, T: Top--Bottom), and texts. Segmentation and orientation are within-subjects factors, while text is a between-subjects factor: subjects see all factor levels in each task, but not all combinations since processing the same text twice induces repetition priming \citep{FrancisSaenz2007}.

\subsection{Tasks}
\label{sec:Exp2Tasks}

We define three experimental tasks: text reproduction (\Copy), error identification (\Scan), and revision (\Revise). We measure speed and accuracy in each task, and minimise ambiguity in the latter by means of contrastive evaluation \citep{Sennrich2017} in the \Scan and \Revise tasks: experimental items are manipulated by inserting an artificial error, and the binary response variable encodes whether or not subjects identify (\Scan) or correct (\Revise) the error.

\subsubsection{Text Reproduction (\Copy)}
\label{sec:Exp2Copy}

In cases where no TM or MT suggestions are available, translators read source text and produce target text. These activities are interleaved \citep{Ruiz2008,Dragsted2010}, and we want to assess how interleaved reading and writing is affected by text presentation. However, this process involves comprehension, and to avoid that subjects will spend time on source text comprehension and target text generation problems -- which are difficult to control for as they will vary among participants and texts -- we ask subjects to copy source text into the target text box(es) of our experimental UIs. As such, this task relates to the technical effort in translation from scratch (\ie, overall effort minus time spent on problem solving). We enforce manual typing by suppressing the use of copy and paste commands, and measure the time it takes subjects to type out entire texts. We calculate accuracy as the number of mistyped characters per text (Levenshtein distance).

\subsubsection{Error Identification (\Scan)}
\label{sec:Exp2Scan}

Translators increasingly work with suggestions from TMs or MT systems \citep{doCarmoMoorkens2020}, which involves target text comprehension: translators scan translation suggestions to decide whether they can be used as-is or need adjustment. Special care must be taken when working with suggestions from neural MT systems as they may read fluently, but contain omitted, added, or mistranslated words \citep{Castilho2017Comparative,Castilho2018Massive}. In the \Scan task, we are interested in whether text presentation impacts the speed and accuracy with which translators can identify such mistakes, which we simulate for better measurability (\Section{Exp2BackgroundPerformance}): we either repeat (Addition) or delete (Omission) word sequences in translations produced by professional translators, insert nonsensical sentences (Wrong Meaning), or leave them unchanged (No Error). Examples are shown in \Table{Exp2ScanExamples}. We apply these manipulations to 10\percent of randomly selected sentences (minimum: 1) in each text, roughly corresponding to the distribution of errors in English to German MT \citep{Castilho2018Massive}. Subjects are asked to assign each text to one of the four categories. We measure how much time they need for each judgement, and whether or not they assign the correct category.

\begin{table}
    \centering
    \input{tables/ScanExamples}
    \caption[Examples of target text manipulations in the \Scan task.]{Examples of target text manipulations in the \Scan task. Subjects see the original source $S$ and manipulated target $T_m$; the original target $T_o$ is shown here for the purpose of illustration. In the experiment, the manipulated sentences are embedded in full news articles.}
    \label{tab:Exp2ScanExamples}
\end{table}

\subsubsection{Revision (\Revise)}
\label{sec:Exp2Revise}

Translations are normally revised before being released, and one important aspect in revision is cohesion: making sure that the connection between sentences and/or paragraphs is appropriate \citep{Shih2006}. Since TMs and MT usually operate on isolated sentences, they are prone to suggest sentences with anaphors (such as pronouns) and named entities (such as product names) that are not compatible with surrounding sentences and the document as a whole \citep{Castilho2017SOTA,Mueller2018}. In the \Revise task, we test if UI segmentation and orientation impact the ability and speed of translators to correct such errors. As in the \Scan task, we manipulate professional translations, and insert one error per document: a mistranslated anaphor or named entity. These errors are constructed such that they are not identifiable within single sentences, meaning subjects have to read the entire text or at least the surrounding sentences to notice them (\Table{Exp2ReviseExamples}). Subjects are asked to revise full documents, and are not told that we focus on anaphors and named entities specifically. We classify the revised documents they submit as correct or incorrect solely based on whether the inserted error is corrected. Any other revisions made by subjects are ignored.

\begin{table}
    \centering
    \input{tables/ReviseExamples}
    \caption[Examples of target text manipulations in the \Revise task.]{Examples of target text manipulations in the \Revise task. Subjects see the original source $S$ and manipulated target $T_m$; the original target $T_o$ is shown here for the purpose of illustration. In the experiment, the manipulated passages are embedded in full news articles.}
    \label{tab:Exp2ReviseExamples}
\end{table}

\subsection{Materials}
\label{sec:Exp2Materials}

\subsubsection{Texts}
\label{sec:Exp2MaterialsTexts}

We use German translations of English news articles in all tasks.
Both the original English texts and their German translations, produced by professional translators, stem from reference data released by the organisers of the 2017 and 2018 Conference on Machine Translation \citep{WMT2017,WMT2018}.\footnote{\url{http://data.statmt.org/wmt17/translation-task/test.tgz}}\textsuperscript{,}\footnote{\url{http://data.statmt.org/wmt18/translation-task/test.tgz}}
Texts are chosen at random, excluding very short and very long instances whose lengths differ by more than one standard deviation from the mean number of sentences per text in the entire collection.
The selected texts contain 21.85 sentences on average (min=8, max=44, median=20.00, sd{=}9.91).
We note that the overall quality of the German translations, which we manipulate by inserting specific errors for the \Scan and \Revise tasks, has been criticised \citep{Hassan2018}; we do not edit or control for errors other than the ones we insert for contrastive evaluation. This is potentially problematic for the \Scan task, \eg, if a translation into which we artificially insert an addition also contains an omission produced by the original translator; in the \Revise task, additional errors cannot influence our measurements since we ignore edits other than those made to our manipulations, and in the \Copy task, subjects only work with the source texts (see above).

\subsubsection{User Interfaces (UIs)}
\label{sec:Exp2MaterialsUIs}

We test four experimental UIs that differ in text presentation, resulting from crossing the levels of two experimental factors: segmentation and orientation.
Sentence-level UIs show sentences in individual text boxes, akin to most CAT tools currently used by professional translators; document-level UIs show the full text in a single text box, as seen in regular word processors.
Interfaces with left--right orientation place target text to the right of the corresponding source text, while target text is placed underneath the corresponding source text in UIs with top--bottom orientation. 
A screenshot of each UI is shown in \Figure{Exp2Screen}.

All UIs use a fixed-width content pane of 1024 by 576 pixels, a 16:9 ratio (the white area in Figures~\ref{fig:Exp2Screen}a--d). Working with texts exceeding the height of this pane requires vertical scrolling. The scrolling behaviour differs between sentence- and document-level UIs in that the former use a single scroll bar (\eg, \Figure{Exp2ScreenST}), while document-level UIs use individual scroll bars for the source and target text boxes (\eg, \Figure{Exp2ScreenDT}). This feature 
is criticised by a number of subjects in our post-experiment survey, as discussed further in \Section{Exp2Limitations}.

Typographic choices are based on design guidelines for on-screen readability \citep{Rello2016,Miniukovich2017}. We use a 12~pt sans-serif font (Arial) to typeset source and target text in dark grey and black colour, respectively, with 150\percent line spacing, left justification, and ragged right edge.

\subsection{Subjects}
\label{sec:Exp2Subjects}

We recruit 20 professional English to German translators (S1--20) from a multinational language services provider, excluding individuals who have participated in our interviews (I1--8, \Section{Exp2Survey}). We pay each translator \$\,245.00 for completing the entire experiment. With an average duration of 7.55 hours, this corresponds to an hourly rate of \$\,32.45, close to the industry average of \$\,35.57.\footnote{\label{ftn:ProZ}According to rates reported by freelance translators and translation companies on ProZ, a large online translation community:  \url{https://search.proz.com/employers/rates?source_lang=eng&target_lang=deu&disc_spec_id=&currency=usd}.}

\begin{table}
    \centering
    \input{tables/Surveys}
    \caption{Background information by number of subjects in the pre-experimental survey.}
    \label{tab:Exp2Surveys}
\end{table}

Subjects complete a pre-experiment survey in which we elicit information on their personal background. On average, subjects have 10.70 years of professional translation experience (min=1, max=28, median=6.50, sd=9.12). 12 out of 20 subjects have a university degree in translation, and 10 have some background in information technology, mostly from attending specialised courses, seminars, or workshops (\Table{Exp2SurveysEducation}). This distribution is very similar to that of a larger group of professional translators surveyed by \citet{Zaretskaya2015}, whereas the percentage of regular or occasional users of translation technology is higher among our subjects (\Table{Exp2SurveysUse}).

\subsection{Procedure}
\label{sec:Exp2Procedure}

As most professional freelance translators work from home \citep{EhrensbergerDow2016}, we opt for a browser-based remote experiment. Subjects complete the experiment using their own computer from a workplace of choice. We block access to the experiment with unsupported browsers, mobile devices, or screens with a resolution that is not large enough to accommodate the entire content pane (see above). We send out general instructions via email, and have a 15-minute introductory call to clarify questions with each subject.

We randomly choose 12, 32, and 12 texts for the \Copy, \Scan, and \Revise tasks, respectively, plus four texts per task for training. We determine the number of items in a pilot run with two professional translators who do not participate in the final experiment. We include more items in \Scan so as to collect sufficient responses for each type of manipulation (Addition, Omission, Wrong Meaning, and No Error). Each subject is presented with the same texts in each task, but not in the same UIs; to control for order effects, we use a Latin square assignment of texts to UIs, and counterbalance the order of tasks among subjects.

Subjects complete the entire experiment in a single workday. Each task starts with a training phase: subjects read through the task instructions and then complete the four training items, one in each UI (in random order). We use a visual cue to distinguish training from experimental items (\Figure{Exp2ScreenDL}), and subjects can repeat each training phase as often as they wish. They can take breaks between tasks, but must complete all items within a task without breaks and under time pressure: we display an idle timer that is reset upon any keyboard or mouse activity, and trigger automatic submission of the current item if no such activity is recorded within three minutes. Time pressure is common in professional translation \citep{EhrensbergerDow2016} and may increase cognitive function \citep{Campbell1999}, but we avoid a fixed deadline to account for per-subject and per-item variation.

Subjects complete a survey before the experiment, one after each task, and one after the experiment. They can optionally leave free-form feedback.

\subsection{Data Analysis}
\label{sec:Exp2DataAnalysis}

Our response variables are time and accuracy. We measure total wallclock time per item, which we log-transform for statistical analysis as our response time measurements follow a log-normal distribution. The coding of accuracy depends on the task, as described in the following section. We report speed in words per hour for \Copy and \Revise, and in seconds per item for \Scan since subjects can make correct judgements without reading through the entire text, thus biasing normalisation by length. We define a word as 5 source text characters, including spaces \citep{ArifStuerzlinger2009}.

We fit linear and logistic mixed-effects models to our measurements for continuous and categorical response variables, respectively, using the \texttt{lme4} package in R \citep{lme4}. We use a random effects structure with random intercepts for subjects and texts in all of our models \citep{Green2013}, and apply mild a-priori screening in combination with model criticism to detect outliers \citep{BaayenMilin2010}. We check for deviations from homoscedasticity or normality by visual inspection of residual plots and Shapiro-Wilk tests in linear models, and inspect logistic models for overdispersion problems and high error rates \citep{GelmanHill2007}.

\section{Experimental Results}
\label{sec:Exp2Results}

\begin{table}
    \centering
    \makebox[\linewidth][c]{
        \input{tables/ResultsOverall}
    }
    \captionsetup{width=1.44\textwidth}
    \caption{Summary of experimental results. Significance levels are denoted by \p~$p{<}.1$, \pp~$p{<}.05$, \ppp~$p{<}.01$, and \pppp~$p{<}.001$.}
    \label{tab:Exp2Results}
\end{table}

\subsection{Text Reproduction (\Copy)}
\label{sec:Exp2ResultsCopy}

Out of the 240 responses in \Copy, we exclude 2 responses triggered by automatic submission due to subject inactivity (no keyboard or mouse input) for three minutes (\Section{Exp2Procedure}).

\subsubsection{Speed}
\label{sec:Exp2ResultsCopyTime}

A-priori screening removes 8 responses with response times deviating by more than 2.5 standard deviations from the per-text (5) and per-subject (3) medians. We fit a linear mixed-effects model for log-transformed response time with fixed effects for segmentation and orientation, and remove 4 overly influential outliers with large residuals through model criticism.

Likelihood ratio tests find significant effects for both segmentation ($\chi^2(1){=}14.58,\allowbreak p{<}.001$) and orientation ($\chi^2(1){=}5.66,\allowbreak p{<}.05$): sentence-level is faster than document-level segmentation, and top--bottom is faster than left--right orientation. A model with an interaction term for segmentation and orientation does not improve model selection scores (\ie, the Akaike (AIC) and Bayesian (BIC) information criteria), and the interaction is not significant ($\chi^2(1){=}0.94, p{=}.33$).

\subsubsection{Accuracy}
\label{sec:Exp2ResultsCopyAccuracy}

We remove 6 responses that contain between 328 and 2230 mistyped characters, more than 2.5 standard deviations from the global mean, which indicates rashness or unintentional submission before completion. Another 6 observations with large residuals are removed through model criticism.

Predicting the number of mistyped characters from segmentation and orientation results in heteroscedastic residuals, so we apply a sqrt-transformation to the dependent variable; a log-transformation is not applicable since 0 -- no typing errors at all -- is a valid response. Likelihood ratio tests find no significant effects for segmentation ($\chi^2(1){=}0.59, p{=}0.44$) and orientation ($\chi^2(1){=}2.08, p{=}0.15$), even with more complex models that include additional predictors such as translator experience or familiarity with translation technology.

\subsection{Error Identification (\Scan)}
\label{sec:Exp2ResultsScan}

\subsubsection{Speed}
\label{sec:Exp2ResultsScanTime}

We are interested in whether text presentation influences the time needed to make correct judgements of target text quality, so we consider correctly labelled (433 out of 640) responses for the \Scan time model. We remove 4 responses with response times below 1 second, and fit a linear mixed-effects model with fixed effects for segmentation and orientation. 16 responses are removed through model criticism. A stepwise variable selection procedure results in a model with better AIC (930.9 \vs 956.6) and BIC (975.2 \vs 980.8) scores, which includes two additional fixed effects: type of manipulation and experience with MT. The latter is elicited in the pre-experiment survey, where we ask subjects if they have used MT \enquote{regularly}, \enquote{sometimes}, or \enquote{never} (\Section{Exp2Subjects}).

Subjects are significantly faster with sentence-level segmentation ($\chi^2(1){=}7.34,\allowbreak p{<}.01$). Likelihood ratio tests do not find a significant effect for orientation ($\chi^2(1){=}0.11, p{=}.74$), but for type of manipulation ($\chi^2(3){=}33.10, p{<}.001$) and experience with MT ($\chi^2(2){=}\allowbreak7.71,\allowbreak p{<}.05$). Subjects detect translations with Wrong Meaning quickly, and need much longer to identify translations that are not manipulated (No Error). This is not surprising since making sure that a translation contains no errors requires that it be read to the end, while subjects can stop reading as soon as they find an error in a manipulated translation. In terms of MT, subjects who have used the technology regularly are the slowest, but also the most accurate.


\subsubsection{Accuracy}
\label{sec:Exp2ResultsScanAccuracy}

We use a binary coding of 1 (when subjects assign the correct class) and 0 (otherwise) for the response variable in \Scan accuracy. After again removing the 4 outliers with response times below 1 second, we fit a logistic mixed-effects model using the same mixed-effects structure as in the time model.

Likelihood ratio tests find no significant effects for segmentation ($\chi^2(1){=}0.11, p{=}.74$) and orientation ($\chi^2(1){=}1.41,\allowbreak p{=}.24$), but for type of manipulation ($\chi^2(3){=}60.67,\allowbreak p{<}.001$) and experience with MT ($\chi^2(3){=}12.98, p{<}.01$). Subjects label 89.2\percent of translations with wrong meaning correctly, more so than translations with missing words (61.6\percent), repeated words (62.2\percent), and translations that contain no error (56.9\percent). This indicates that translations we have not manipulated contain errors that we do not control for (\Section{Exp2MaterialsTexts}). As for experience with MT, 80.0\percent of responses produced by subjects who have regularly used the technology are correct, more than those of subjects who have sometimes (65.7\percent) or never (58.2\percent) worked with MT.


\subsection{Revision (\Revise)}
\label{sec:Exp2ResultsRevise}

We build separate models for texts manipulated with a wrong named entity and a wrong anaphor, and remove one response each where no keyboard or mouse activity was recorded for more than three minutes (\Section{Exp2Procedure}).

\subsubsection{Speed}
\label{sec:Exp2ResultsReviseTime}

As in \Scan, we model response time for texts which subjects revised correctly. A-priori screening removes 5 responses (2 with a manipulated named entity, 3 with a manipulated anaphor) with response times deviating by more than 2.5 standard deviations from the per-text median, leaving a total of 63 and 57 correct responses for texts with a manipulated named entity and anaphora, respectively.

We find a significant effect for orientation with texts containing a manipulated named entity ($\chi^2(1){=}6.30, p{<}.05$), which subjects revise faster with left--right UIs. Effects for segmentation with these texts ($\chi^2(1){=}0.08, p{=}.77$) as well as both segmentation ($\chi^2(1){=}0.01, p{=}.91$) and orientation ($\chi^2(1){=}0.15, p{=}.70$) with texts containing a manipulated anaphor are not significant.



\subsubsection{Accuracy}
\label{sec:Exp2ResultsReviseAccuracy}

We use a binary coding of 1 (when subjects correct the error we deliberately inserted) or 0 (when they do not) in the response variable for accuracy. We fit a logistic mixed-effects model each to the responses from texts with a manipulated named entity and anaphor, excluding one text with a named entity that none of the subjects revise correctly (20 responses). We include experience with MT as a fixed effect in addition to segmentation and orientation, which improves model selection scores for the named entity model and leads to lower error rates \citep{GelmanHill2007} in both models.

For texts with a manipulated named entity, effects for segmentation ($\chi^2(1){=}0.00, p{=}.94$) and orientation ($\chi^2(1){=}0.61, p{=}.44$) are not significant, but revision by regular users of machine translation (experience with MT) is significantly less accurate ($\chi^2(2){=}6.55,\allowbreak p{<}.05$).

For texts with a manipulated anaphor, likelihood-ratio tests find a near-significant effect of segmentation ($\chi^2(1){=}5.53, p{=}.06$), where mean accuracy is higher with document-level UIs. Effects for orientation ($\chi^2(1){=}0.26, p{=}.61$) and experience with MT ($\chi^2(2){=}1.28,\allowbreak p{=}.53$) are not significant.



\subsection{UI Preference}
\label{sec:Exp2Surveys}

\begin{figure}
    \centering
    \makebox[\linewidth][c]{
        \includegraphics[width=1.44\textwidth]{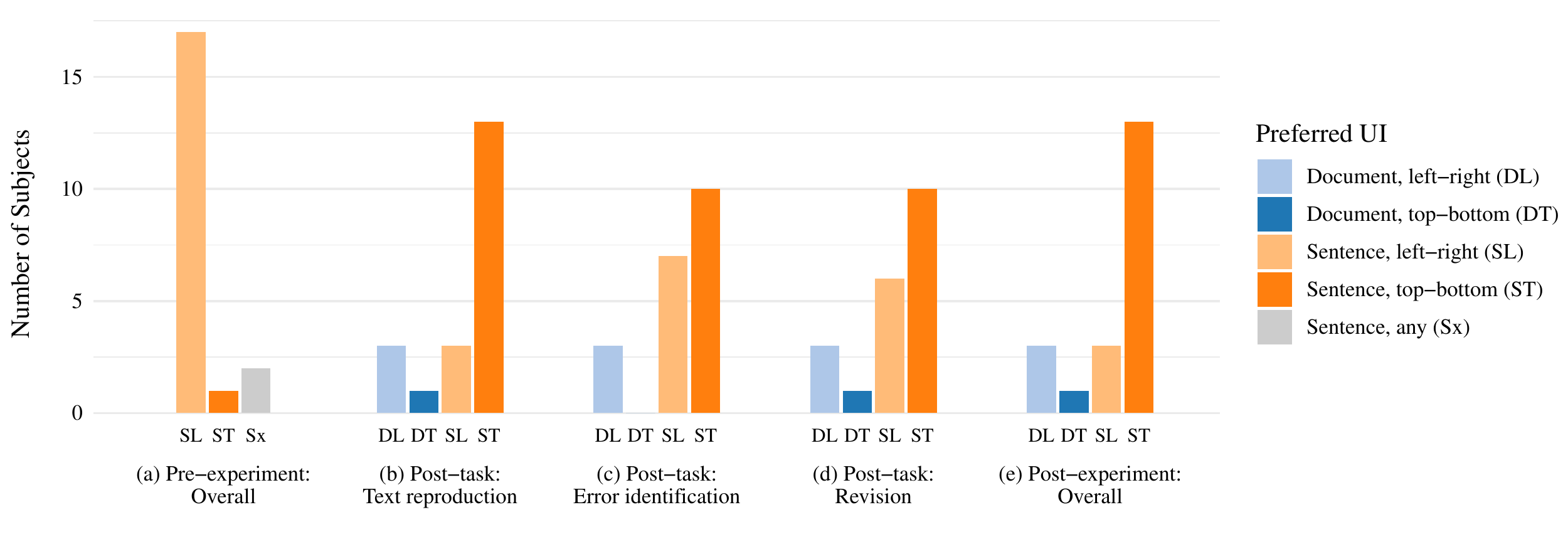}
    }
    \captionsetup{width=1.44\textwidth}
    \caption[UI preferences.]{UI Preferences. The majority of subjects are used to sentence-level UIs with left--right orientation (a), but prefer top--bottom orientation in the experiment (e).}
    \label{fig:Exp2UIPreferences}
\end{figure}

We ask subjects about their preferred orientation (left--right or top--bottom) in the CAT tools they usually work with in the pre-experiment survey, and contrast this with feedback on the experimental UIs in three post-task surveys (one each after \Copy, \Scan, and \Revise) and a post-experiment survey. As shown in \Figure{Exp2UIPreferences}, most subjects (85\percent) use left--right orientation in their daily work. In the experiment, however, the majority (65\percent) prefer the sentence-level UI with top--bottom orientation overall, according to the post-experiment survey.

\section{Discussion and Design Implications}
\label{sec:Exp2Discussion}

The triangulation of preliminary feedback from potential users (I1--8, \Section{Exp2Survey}), our empirical results, and feedback from experimental subjects (S1--20) yields new design principles for text presentation in CAT tools. We also discuss the limitations of our study in this section, and outline directions for future work.




\subsection{Segmentation}

CAT tools in wide use visually separate the sentences in a document. Our results suggest that this is helpful for sentence-level tasks: reproducing text (\Copy) and identifying errors within sentences (\Scan) are significantly faster in UIs with sentence-by-sentence presentation (\Table{Exp2Results}). In post-experiment feedback, several subjects note that it is \enquote{easier to work with a text when it is separated into segments} (S4), notably because it \enquote{eliminates [the] need for scrolling and paragraphing} (S1).

In the \Revise task that requires super-sentential context, on the other hand, segment-by-segment presentation provides no advantage over a display of continuous text. On the contrary, anaphoric relations are revised more accurately in document-level UIs, and while the difference to sentence-level UIs is not significant ($p{=}.06$), the effect size is considerable: 58.07\percent in DT \vs 37.93\percent in ST, the latter performing best in \Copy and \Scan (\Table{Exp2Results}).

In this light, the characterisation of sentence-by-sentence presentation as \enquote{unnatural} \citep{Dragsted2006} or \enquote{irritating} \citep{OBrien2017} in other translation research and the largely positive feedback from our subjects are not necessarily conflicting: the suitability of sentence segmentation depends on the task, and the problem may be that it cannot be turned off when it is not suitable. Since the translation activities that motivate our experimental tasks are interleaved in practice \citep{Ruiz2008,Dragsted2010}, letting translators switch between a segmented and continuous view of the document they are translating may enable them to focus on local context and consider global context when needed. This is supported by feedback from our interviewees, who mention that \enquote{you need segmentation sometimes to just better focus} (I5), and suggest that combining sentence-by-sentence with full document presentation would be most effective (I2, I5, I7).

\subsection{Orientation}

Most of the CAT tools currently available to professional translators display source and target sentences side-by-side (left--right). \citet{Green2014PTM} conjecture that this \enquote{spreadsheet design may not be optimal for reading}, proposing a top--bottom arrangement instead. Our results show that top--bottom orientation can indeed be helpful, but not for reading: it significantly accelerates text reproduction (\Copy), but provides no advantage over left--right orientation in tasks that involve no (\Scan) or little writing (\Revise). Conversely, left--right orientation enables faster revision, significantly so in the Named Entity subtask (\Table{Exp2Results}).

Top--bottom orientation is surprisingly popular among subjects. S9 comments that \enquote{I first thought it might be weird to work like this without having a continuous source text to look at (because the source text is interrupted by the target text), but it worked like a breeze}, and S12 notes that in contrast to what they are used to (left--right), \enquote{the orientation is a bit different, but it doesn't bother me}. S19 remarks that \enquote{I have never arranged my programs this way and I might have to}. While only one subject states they prefer the sentence-level UI with top--bottom orientation (ST) before the experiment, 13 out of 20 subjects prefer it thereafter. For reading-intensive tasks, however, it is preferred less often that for the writing-intensive \Copy task (\Figure{Exp2UIPreferences}).

As such, our results motivate the use of top--bottom orientation in UIs to support writing. Revision, on the other hand, is faster with left--right orientation (\Table{Exp2Results}).

\subsection{Limitations}
\label{sec:Exp2Limitations}

As motivated in \Section{Exp2BackgroundPerformance}, our experimental design maximises internal validity: subjects perform tasks reflecting specific activities that can be relevant when working with CAT tools, but these tasks do not mirror real-life translation in which many activities are interleaved \citep{Ruiz2008,Dragsted2010}. Since some of the UIs we explore are not available in widely used CAT tools, conducting an experiment under realistic working conditions has not been an option for our investigation. Our experimental design allows more fine-grained conclusions instead: for example, we can empirically show that the suitability of sentence segmentation is task-dependent, suggesting that the segment-by-segment presentation in CAT tools should not be replaced, but complemented with an unsegmented view of the text.\footnote{We note that some CAT tools offer a feature to visualise the translated segments in a preview of the target document, but this preview is not editable.} Nevertheless, our findings are pending confirmation in real-life settings, for which our experimental UIs should be incorporated into fully functional CAT tools.

Post-experiment feedback on our prototypical UIs may inform this transition. In particular, the scrolling behaviour in document-level UIs turns out to be more important than we anticipated. In our experimental UIs, the source and target text panes have separate scroll bars, and scrolling in one of them does not automatically invoke scrolling in the other. The reason is that since source and target texts may differ in length, distance-based synchronisation -- \ie, if the user scrolls down two lines in the source text, also scroll down two lines in the target text -- may result in incorrect alignment. Post-experiment feedback suggests that \enquote{linked scrolling of the two panes \ldots would greatly improve productivity} (S18), particularly with the document-level UI that uses a top--bottom arrangement of source and target documents, which S14 called \enquote{a scrolling and matching nightmare}. Surprisingly, the fastest and most accurate results in one of the revision subtasks (Anaphor) are achieved with document-level UIs despite this shortcoming (\Table{Exp2Results}, DL and DT), and 4 out of 20 subjects state that they find one of these UIs most suitable for the experimental tasks overall (\Figure{Exp2UIPreferences}e).

Lastly, our full-day experiment concentrates on a single language pair (English to German) and domain (news), and involves 22 professional translators (20 plus 2 for a pilot run). Many studies use students or crowd workers instead \citep[\eg,][]{Bowker2005,Karimova2018}, and involve a smaller number of subjects and experimental items \citep[\eg,][]{Macklovitch2006,Coppers2018}. Nevertheless, we acknowledge that involving further languages, domains, and more subjects would strengthen our results.

\subsection{Future Work}
\label{sec:Exp2FutureWork}

In future work, our experimental UIs should be integrated into richer prototypes or, ideally, a fully fledged CAT tool. Testing the impact of text presentation on translation under realistic working conditions will require many features that are not available in our experimental prototypes, starting with real-time integration of translation suggestions from TMs and/or MT. We have investigated four UIs in three experimental tasks, and our results motivate two avenues for further research in particular: the replacement of left--right with top--bottom orientation in sentence-level interfaces, and the use of document-level interfaces for revision. We consider the latter to be important since the quality of machine-generated translation suggestions is improving steadily \citep[\eg][]{JunczysDowmunt2019}, which may reduce the amount of writing needed to produce publication-quality translations and in turn increase the need for UIs that are optimised for revision.

Our study also sheds light on how experience with MT affects accuracy in professional translation. Regular users of MT detect significantly more errors within sentences than occasional or non-users (\Section{Exp2ResultsScanAccuracy}), but are the least accurate in revising incoherently translated named entities across sentences (\Section{Exp2ResultsReviseAccuracy}). Future work will have to investigate whether the strong focus on single sentences in MT system outputs -- and/or in the UI layout of CAT tools -- has a priming effect on professional translators.

\section{Conclusions}
\label{sec:Exp2Conclusions}




In a controlled experiment with 20 professional translators, we tested the impact of changes in text presentation on speed and accuracy in three text processing tasks. We found that:

\begin{itemize}
    \item Sentence-by-sentence presentation enabled faster text reproduction (\Copy) and within-sentence error identification (\Scan) compared to unsegmented text; it did not enable faster revision (\Revise).
    \item Presentation of documents (unsegmented text) lead to the highest accuracy in revision for anaphoric relations between sentences (\Revise, Anaphor).
    \item Top--bottom orientation of source and target sentences enabled faster text reproduction (\Copy) than left--right orientation, and was preferred by the majority of subjects in all experimental tasks.
    \item Left--right orientation enabled faster revision for lexical cohesion (\Revise, Named Entity).
\end{itemize}

Our results suggest that the impact of text presentation has been overlooked in the conception of translation technology. Widely used CAT tools implement sentence-by-sentence presentation with left--right orientation for both translation and revision, but our measurements and feedback from subjects imply that source and target sentences should be presented in a top--bottom arrangement, and that CAT tools should offer a side-by-side view of unsegmented text for revision. Most commercial systems do not support these UI layouts, and should be revisited as, at least within the scope of our controlled experiment, they have have a significant impact on translator performance.

\bibliographystyle{acl}
\bibliography{references}

\end{document}

%% file: commands.tex
\newcommand{\Section}[1]{Section~\ref{sec:#1}}
\newcommand{\Table}[1]{Table~\ref{tab:#1}}
\newcommand{\Figure}[1]{Figure~\ref{fig:#1}}

\newcommand{\ie}{i.e.\xspace}

\newcommand{\eg}{e.g.\xspace}

\newcommand{\ibid}{\textit{ibid.}\xspace}

\newcommand{\percent}{\,\%\xspace}
\newcommand{\etc}{\ldots\xspace}
\newcommand{\sic}{[sic]\xspace}
\newcommand{\vs}{\textit{vs.}\ }
\newcommand{\p}{$\circ$\xspace} 
\newcommand{\pp}{$\bullet$\xspace} 
\newcommand{\ppp}{$\bullet\,\bullet$\xspace} 
\newcommand{\pppp}{$\bullet\bullet\bullet$\xspace} 

\newcommand{\citeg}[1]{\citeauthor{#1}'s \citeyearpar{#1}} 

\newcommand{\Copy}{\textsc{Copy}\xspace}

\newcommand{\Scan}{\textsc{Scan}\xspace}

\newcommand{\Revise}{\textsc{Revise}\xspace}

%% file: tables/ScanExamples.tex
\small
\raggedright

\begin{subtable}{\textwidth}
    \begin{tabularx}{\textwidth}{p{2mm}X}
        \toprule
        $S$    & While sufferers are usually advised to dodge meat and dairy to soothe their symptoms, researchers at Washington University found protein's essential amino acid tryptophan helps develop immune cells that foster a tolerant gut.\\
        $T_o$  & Während den Betroffenen normalerweise geraten wird, Fleisch und Milchprodukte zu meiden, um ihre Symptome zu lindern, fanden Forscher an der Washingtoner Universität heraus, dass die essentielle Aminosäure Tryptophan von Proteinen dazu beiträgt, Immunzellen zu entwickeln, die einen toleranten Darm fördern. \\
        $T_m$  & Während den Betroffenen normalerweise geraten wird, Fleisch und Milchprodukte zu meiden, um ihre Symptome zu lindern zu lindern zu lindern, fanden Forscher an der Washingtoner Universität heraus, dass die essentielle Aminosäure Tryptophan von Proteinen dazu beiträgt, Immunzellen zu entwickeln, die einen toleranten Darm fördern. \\
        \bottomrule
    \end{tabularx}
    \caption{Addition}
    \label{tab:Exp2ScanExamplesAddition}
    \vspace{1em}
\end{subtable}

\begin{subtable}{\textwidth}
    \begin{tabularx}{\textwidth}{p{3mm}X}
        \toprule
        $S$    & Patrick Roy resigned as coach and vice president of the hockey operations of the Colorado Avalanche on Thursday, citing a lack of a voice within the team's decision-making process. \\
        $T_o$  & Patrick Roy trat am Donnerstag als Trainer und Vice President Of Hockey Operations der Colorado Avalanche zurück und führte ein zu geringes Mitbestimmungsrecht beim Entscheidungsprozess des Teams an. \\
        $T_m$  & Patrick Roy trat am Donnerstag als Trainer und Vice President Of Hockey Operations zurück und führte ein zu geringes Mitbestimmungsrecht beim Entscheidungsprozess des Teams an. \\
        \bottomrule
    \end{tabularx}
    \caption{Omission}
    \label{tab:Exp2ScanExamplesOmission}
    \vspace{1em}
\end{subtable}

\begin{subtable}{\textwidth}
    \begin{tabularx}{\textwidth}{p{3mm}X}
        \toprule
        $S$    & Heavy rain, flooding prompts rescues in Louisiana, Mississippi \\
        $T_o$  & Heftige Regenfälle, Überschwemmung gibt Anlass zu Rettungen in Louisiana, Mississippi \\
        $T_m$  & Öffnen Sie im zweiten Fenster das eigene Persönliche Verzeichnis. \\
        \bottomrule
    \end{tabularx}
    \caption{Wrong Meaning}
    \label{tab:Exp2ScanExamplesWrongMeaning}
    \vspace{1em}
\end{subtable}

\begin{subtable}{\textwidth}
    \begin{tabularx}{\textwidth}{p{3mm}X}
        \toprule
        $S$    & The ``Made in America'' event was designated an official event by the White House, and would not have been covered by the Hatch Act. \\
        $T_o$  & Die Veranstaltung ``Made in America'' wurde vom Weißen Haus als offizielle Veranstaltung bezeichnet und wäre nicht vom Hatch Act abgedeckt worden. \\
        $T_m$  & Die Veranstaltung ``Made in America'' wurde vom Weißen Haus als offizielle Veranstaltung bezeichnet und wäre nicht vom Hatch Act abgedeckt worden. \\
        \bottomrule
    \end{tabularx}
    \caption{No Error}
    \label{tab:Exp2ScanExamplesNoError}
    \vspace{1em}
\end{subtable}

%% file: tables/ReviseExamples.tex
\small
\raggedright

\begin{subtable}{\textwidth}
    \begin{tabularx}{\textwidth}{p{3mm}X}
        \toprule
        $S$    & \etc to an Earls Court apartment in 2014. It is on the first floor of a smart Queen Anne terrace - and it is a testament to the new design \etc \\
        $T_o$  & \etc zu einem Earls Court Apartment. Es liegt im ersten Stock einer eleganten Queen Anne Terrasse - und es ist ein Beweis für das neue Design \etc \\
        $T_m$  & \etc zu einem Earls Court Apartment. Sie liegt im ersten Stock einer eleganten Queen Anne Terrasse - und es ist ein Beweis für das neue Design \etc \\
        \bottomrule
    \end{tabularx}
    \caption{Anaphor}
    \label{tab:Exp2ReviseExamplesAnaphora}
    \vspace{1em}
\end{subtable}

\begin{subtable}{\textwidth}
    \begin{tabularx}{\textwidth}{p{2mm}X}
        \toprule
        $S$    & Pokémon Go, a worthy hunt for health and happiness \etc Within days, Pokémon Go had more users than Tinder \etc But the beauty of Pokémon Go is it gets people outside doing something they enjoy \etc \\
        $T_o$  & Pokémon Go, eine Jagd nach Gesundheit und Glück, die sich lohnt \etc Innerhalb weniger Tage hatte Pokémon Go mehr Benutzer als Tinder \etc Aber das Wundervolle an Pokémon Go ist, dass es die Leute dazu bringt, etwas im Freien zu tun \etc  \\
        $T_m$  & Pokémon Go, eine Jagd nach Gesundheit und Glück, die sich lohnt \etc Innerhalb weniger Tage hatte Pokémon Go mehr Benutzer als Tinder \etc Aber das Wundervolle an Pokémon Gehen ist, dass es die Leute dazu bringt, etwas im Freien zu tun \etc  \\
        \bottomrule
    \end{tabularx}
    \caption{Named Entity}
    \label{tab:Exp2ReviseExamplesNamedEntity}
    \vspace{1em}
\end{subtable}

%% file: tables/Surveys.tex
\centering



\begin{subtable}{\textwidth}
    \centering
    \begin{tabular}{lrrrrr}
    \toprule
                               & None & Courses & Bachelor's & Master's & PhD \\ \midrule
        Translation            & 5    & 3       & 7          & 4        & 1   \\
        Information Technology & 10   & 9       & 0          & 1        & 0   \\
    \bottomrule
    \end{tabular}
    \caption{Education (highest degree)}
    \label{tab:Exp2SurveysEducation}
\end{subtable}

\bigskip

\begin{subtable}{\textwidth}
    \centering
    \begin{tabular}{lrrr}
    \toprule
                               & Regular & Sometimes & Never \\ \midrule
        Translation Memory     & 15      & 4         & 1     \\
        Terminology Management & 7       & 9         & 4     \\
        Machine Translation    & 5       & 10        & 5     \\
        Quality Assurance      & 4       & 8         & 8     \\
    \bottomrule
    \end{tabular}
    \caption{Use of translation technology}
    \label{tab:Exp2SurveysUse}
\end{subtable}



%% file: tables/ResultsOverall.tex
\small
\renewcommand{\arraystretch}{1.2}
\begin{tabular}{lllllllllllll}
\toprule
Task              && \multicolumn{2}{l}{\Copy} && \multicolumn{2}{l}{\Scan} && \multicolumn{2}{l}{\Revise: Named Entity} && \multicolumn{2}{l}{\Revise: Anaphor} \\[0.5em]
Response Variable && Speed      & Accuracy   && Speed    & Accuracy     && Speed           & Accuracy         && Speed          & Accuracy         \\
Unit              && words/h    & \# typos   && s/item   & \percent correct   && words/h         & \percent correct       && words/h        & \percent correct       \\ 
Sample Size       && 226        & 226        && 413      & 636          && 63              & 99              && 57             & 119              \\[0.5em] 
\multicolumn{9}{l}{\it Mean Response}                                                                                                              \\[0.2em]
Sentence, left--right (SL)               && 2,523.16   & 9.49       && 114.31   & 66.46        && 5,037.84        & 61.54            && 4,693.47       & 48.27            \\
Sentence, top--bottom (ST)               && 2,639.13   & 8.07       && 123.80   & 69.38        && 4,323.27        & 76.00            && 4,511.23       & 37.93            \\
Document, left--right (DL)               && 2,463.37   & 9.19       && 153.98   & 65.19        && 4,786.27        & 70.83            && 5,350.81       & 56.67            \\
Document, top--bottom (DT)               && 2,459.79   & 9.28       && 145.83   & 68.75        && 4,365.75        & 66.67            && 4,778.32       & 58.07            \\[0.5em] 
\multicolumn{9}{l}{\it Effects}                                                                                                                    \\[0.2em]
Segmentation      && \pppp &          && \ppp  &          &&       &       &&       & \p    \\
Orientation       && \pp   &          &&       &          && \ppp  &       &&       &       \\
Experience with MT && $n/a$ & $n/a$   && \ppp  & \pppp    && $n/a$ & \pp   && $n/a$ &       \\
Type of Manipulation && $n/a$ & $n/a$    && \pppp & \pppp    && $n/a$ & $n/a$ && $n/a$ & $n/a$ \\
\bottomrule
\end{tabular}